\documentclass[15pt]{article}
\usepackage{graphicx}
\begin{document}
\addtolength{\textwidth}{2.75in}
\addtolength{\topmargin}{-0.875in}
\addtolength{\textheight}{1.75in}
\title{{\bf A note on the physical interpretation of neural PDE's}}
\author{Sauro Succi\\
Italian Institute of Technology,\\ 
Viale Regina Elena, 291, 00161, Rome, Italy\\
CNR-IAC, via dei Taurini, 19, 00185, Roma, Italy\\
Physics Department, Harvard University, Cambridge USA
}
\maketitle

{\bf Abstract}

\vspace{5mm}

We highlight a formal and substantial analogy between Machine Learning (ML) 
algorithms and discrete dynamical systems (DDS) in relaxation form.
The analogy offers a transparent interpretation of the 
weights in terms of physical information-propagation processes and
identifies the model function of the forward ML
step with the local attractor of the corresponding discrete dynamics.
Besides improving the explainability of current ML applications, this
analogy may also facilitate the development of a new class 
ML algorithms with a reduced number of weights. 

\section{Introduction}

Machine Learning (ML) has taken science (and society) by storm 
in the last decade, with numerous applications which seem
to defy our best theoretical and modeling tools \cite{ML}.
Leaving aside a significant amount of hype, ML raises
a number of genuine hopes to counter some the most 
vexing challenges for the scientific method, particularly
the curse of dimensionality \cite{CoD}.
This however does not come for free; in particular the current
trends towards the use of an astronomical number of parameters
(trillions in the case of recent chatbots), none of which lends 
itself to a direct physical interpretation, jointly with an unsustainable
power demand, beg for a change of strategy, namely less weights and
more insight \cite{CHAT}.
In this paper, we present an attempt along this line.
In particular, by highlighting the one-to-one mapping
between ML procedures and discrete dynamical systems, we
suggest that ML could possibly be conducted by means
of a restricted and more economical class
of weight matrices, each of which can be interpreted as
a specific information-propagation process.     

\section{The basic ML procedure}

The basic idea of ML is to represent a given $d$-dimensional 
output $y_T$ (target) through the recursive application 
of a simple nonlinear and nonlocal map \cite{ML}.
For a neural network (NN) consisting of an input layer $x$,  $L$ hidden layers
$z_1 \dots z_L$, each containing the same number of $N$ neurons (Transformers), and 
an output layer $y$,
the update chain $x \to z_1 \dots \to z_L \to  y$
reads symbolically as follows:
\begin{eqnarray}
\label{ML}
z_0 = x\\
z_1 = f_1(W_1 z_0       -b_1),\; \dots  z_L = f_L(W_L z_{L-1} -b_L),\\
z_{L+1} = f_{L+1}(W_{L+1} z_L -b_{L+1})=y
\end{eqnarray}
where $W_l$ are $N \times N$ matrices of weights, $b_l$ are N-dimensional arrays
of biases and $f$ is a nonlinear activation function, to be chosen out 
of a large palette of possibilities.
The output $y$ is then compared with a given training 
target $y_T$ (Truth) and the weights are
recursively updated in such a way as to minimize 
the discrepancy between $y$ and $y_T$ (Loss function), up 
to the desired tolerance, the so-called "back-propagation" step \cite{BACK}.  
In equations:
\begin{equation}
\label{LOSS}
\mathcal{L}[W] = dis (y[W],y_T) \le \epsilon
\end{equation}
where the loss function $\mathcal{L}$, namely the distance between
the model output and the target in some suitable metric, is
an explicit function of the weights configuration $W$.
The central engine of the ML procedure is an update schedule
of the weights whereby the loss function is taken below 
the desired tolerance $\epsilon$. 
This is typically obtained through some form of 
steepest descent (SD) search:

\begin{equation}
\label{SD}
W(\tau+d \tau)=W(\tau)- \alpha \frac{\partial \mathcal{L}}{\partial W}
\end{equation}

where $\tau$ denotes the iteration step and 
$\alpha$ is a suitable relaxation rate.

The idea is that with {\it big enough} data for training, 
the ML sequence (1-2-3) can reach {\it any} target, whence 
the alleged demise of the scientific method \cite{WIRED,PEDRO}.

Where does such magic come from?

The key point is that for a DNN (deep neural net) of depth $L$
(number of layers) and width $N$ (numbers of neurons
per layer), there are $N_P=N^L$ possible paths connecting any 
single item $x_i, \; i=1,N$ in the input layer to any another 
single item $y_j, \;j=1,N$ in the output layer.
Hence a DNN with $N=10^3$ neurons and $L=10^2$ layers features
$N_W = N^2 L=10^8$ weights and $N_P=10^{30}$ paths.
Such gargantuan network of paths represents the state-space
of the DNN learning process, which can proceed through several
concurrent paths at a time. 
These numbers unveil the magic behind ML: DNN duel the 
CoD face up, by unleashing an exponential number of paths, and
adjusting them in such a way as to sensibly populate the 
sneaky regions where the golden nuggets are to be found.
It would far too simplistic to accomodate the success of ML just
in terms of this exponential capacity, the story is subtler than that.
For instance, exponential capacity per-se does not necessarily
explain the "magical" property of sidestepping overfitting, as often observed
in large scale ML applications, typically Large Language Models \cite{GOOGLE}. 
The ML strategy described above is an opinion splitter: AI pragmatists 
are enthusiastic at the conceptual simplicity of this black-box and, 
speaking of weights, to them  "too much is not enough". 
Scientists fond of Insight ahead of Control, are 
dismayed at the diverging number of parameters, 
their "prejudice" being that parameters are fudge factors
concealing lack of understanding, so that their motto
is rather "the least the best". 
Insight is basically Extreme Underfitting: Newton's law captures
the physics of {\it any} classical gravitational system by means 
of a single parameter, the gravitational constant $G$!
The triumph of insight. 
Last but not least, current leading edge ML applications, such  
as Large Language Models motoring the most powerful "ask-me-anything" chatbots
employ hundreds of billions weights, basically the number
of neurons in our brain, except that our brain works at 20 Watt 
while the largest ML models are now sucking up at least ten million times more. 

In view of the above, there is a recognized need
for a better and deeper theory of machine learning. 
A promising direction in this respect consists in highlight and possibly exploiting the
substantial analogies between DNN and DDS \cite{E1,E2,E3},
and the development of so called "neural" ODE's and PDE's.  

In this paper we offer a potentially new angle along this direction,
by interpreting the neural weights as discrete realizations
of an integral kernel serving as a linear PDE-generator, whose 
action can be given a transparent interpretation in terms of 
information dynamics.
The linear PDE's is then passed to the nonlinear activation function which "scrambles" the 
various linear non-local terms associated with the aforementioned kernel, giving
rise to a corresponding nonlinear neural PDE.

\section{Transformers as continuum dynamical systems}

In this paper we shall refer to "transformers", namely neural networks with
a constant number of weights across the layers.
For the sake of physical concreteness, let us consider a physical interface 
whose altitude $h$ is described by a one-dimensional function
\begin{equation}
h = z(q,t)
\end{equation}
where $q$ is a generalized position and $t$ is a (possibly fictitious)
time coordinate.
The interface altitude is assumed to obey the following 
first order non-linear and non-local dynamics:
\begin{equation}
\label{LIU}
\partial_t z = -\gamma \mathcal{L}(z)\\
\end{equation}
where $\mathcal{L}$ is the (generally nonlinear) Liouville operator
and $\gamma$ is a relaxation frequency.
Let us further assume the Liouvellean admits a non-zero local equilibrium
(non trivial vacuum) obeying 
$$
\mathcal{L}(z^{eq}) = 0
$$
Under such an assumption, we can always recast the Liouville
dynamics in relaxation form:
\begin{equation}
\label{RELAX}
\partial_t z = -\gamma (z-z^{eq})
\end{equation}
where, by construction, the local equilbrium is given by
\begin{equation}
\label{LEQ}
z^{eq}=z -\mathcal{L}(z)
\end{equation}

The next step is consider a neural PDE (NPDE) also in relaxation form

\begin{equation}
\label{NEU}
\partial_t z = -\gamma (z-f[Z(q,t)]),\\
\end{equation}
with initial condition $z(q,t=0)=x(q)$ and normalization 
constraint $||z(t)||=1$.
In the above, $f(Z)$ is a local nonlinear function(al) and 
\begin{equation}
\label{ZETA}
Z(q,t) = \int W(q,q') z(q',t) dq'-b(q,t)
\end{equation}
is the weight-transform of the original signal $z(q,t)$ via
the linear and nonlocal kernel $W(q,q')$, $b(q)$ 
being the bias function.

Direct identification of (\ref{NEU}) with (\ref{RELAX}) delivers 
the expression of the local equilibrium in terms of the activation 
function, namely 
\begin{equation}
\label{EQ1}
z^{eq}(q,t) = z - \mathcal{L}(z) = f[Z(q,t)]
\end{equation}
and, equivalently, the Liouvillean in terms of the activation function
\begin{equation}
\label{EQ2}
\mathcal{L}(z) = z(q,t)- f[Z(q,t)]
\end{equation}

The equivalence of the evolution problem (\ref{NEU},\ref{ZETA}) and the
forward step of the ML procedure is readily exposed by marching
the scheme in time with a forward Euler scheme, as we shall detail shortly.

Here, we note that the evolution consists of the superposition of a 
"slow mode", the local equilibrium $z^{eq}$ 
and a "fast" non-equilibrium component $z^{neq} = z-z^{eq}$,
which relaxes to the equilibrium on a timescale $\gamma^{-1}$.
This relaxation is never complete, at least not 
as long $z^{eq}$ keeps evolving, although on a longer timescale.
Indeed both $z$ and $z^{eq}$ evolve in time until they eventually 
settle down to a steady-state attractor $z^*$, defined by the 
self-consistent condition:
\begin{equation}
z^* = z^{eq,*} = f[W^*z^*-b]
\end{equation}
where indices are suppressed for simplicity.

To be noted that in the course of this evolution the weights 
may also change in time, which is why in the above equation we have
taken $W^* = W(t^*)$.   
The existence and uniqueness of such local attractor(s) may
be useful to inform the optimal depth of the circuit, based on the
value of the Loss function layer by layer. 
Indeed there is no reason why the attractors should coincide
with local minima of the Loss function, hence there is no reason
to wait for the dynamics to reach the attractors instead of trying
to catch the target "on the fly".

\subsection{The weight kernel as a PDE generator}

The neural PDE (\ref{NEU}) is a nonlinear integro-differential equation 
which can be turned into an equivalent family of nonlinear PDE's. 
As anticipated above, the action of the kernel $W(q,q')$ is to turn
the original signal $z(q)$ into the W-transformed signal $Z(q)$.
Next, we show that this transformation can be interpreted in terms 
of generalized advection and diffusion processes.
To this purpose, let us express the linear transform as follows
\begin{equation}
\bar z (q) = \int W(q,q+r) z(q+r) dr 
\end{equation}
where $r=q'-q$.
Assuming that $z(q+r)$ can be expanded in Taylor series, we obtain
\begin{equation}
\bar z(q) = W_0(q) z + W_1(q) z_q + \frac{W_2(q)}{2} z_{qq} + \dots  
\end{equation}
where we have defined 
\begin{equation}
\label{MOME}
W_k(q) = \int r^k W(q,q+r) dr 
\end{equation}
as the $k$-th order moment of the kernel $W(q,q')$ with respect to the displacement $r=q'-q$.
Note that for homogeneus kernels, $W(q,q')=W(r)$, hence all the moments are 
constant, so that $Z(q)$ is the result of a standard convolution, a cornerstone
of image processing \cite{CNN}. 

The inhomogeneous case is obviously much richer.
Clearly the various moments, including their very existence, depend strictly
on the specific form of the weight kernel $W(q,q')$, each class of kernels generating a corresponding
set of nonlinear and nonlocal PDES's.
For instance, homogeneous Gaussian kernels $W(r) \propto e^{-r^2/2 \sigma^2}$, give 
rise to a convergent sequence of moments at all orders, while power-law kernels
support only a finite number of convergent moments, meaning by this 
that such moments diverge with the size of the integration domain.   

The above expression provides a clear clue to Explainable ML, because 
every finite moment $W_k(q)$ carries a concrete physical meaning in 
terms of information propagation processes.
Specifically, $W_0(q)$ is a local amplitude rescaling, $W_1(q)$ is 
local propagation speed and $W_2(q)$ is a local diffusion coefficient.
These three basic processes alone describe a broad spectrum of physical 
phenomena, hence encoding a great deal of expressive power within the
corresponding ML procedure.
Further freedom comes from the higher order moments.
On general grounds, higher moments of even order can be classified as 
generalized diffusion processes, hence they smear out the signal (denoising),
as long as they come with the right sign (positive for $W_2$, negative for $W_4$ and so on). 
Otherwise, they do just the opposite.
For instance, $W_4$, usually called hyper-diffusion, corresponds to a diffusion process
in which the root mean square displacement scales like the power $1/4$ of time.  
Odd higher order moments correspond to generalized propagation; for instance 
$W_3$ is  associated to dispersion processes, namely propagation with a scale-dependent 
velocity, which leads to deformations of the profile $z(q)$. 

PDE's for physical systems rarely go beyond fourth order, but the 
dynamics of information needs not be subjected to such constraint, whence
the scope for more general kernels in ML applications.
Yet, the picture given above still applies, which is a
decide help for the physical interpretation of the weights. 

Once the original signal $z(q)$ is transformed into $\bar z(q)$ 
via the weight kernel, the bias is subtracted to provided the 
input $Z(q) = \bar z(q) - b(q)$ to the local nonlinear
functional which defines the local equilibrium
of the "neural" dynamics:
$$
z^{eq} = f(Z)
$$ 
At this stage, all the generalized advection-diffusion terms described above
are locally "scrambled", thereby injecting a major lease of freedom into 
the class of targets that can be reached by the "neural" dynamics of the system.   

This completes the description of the ML update as an 
information dynamics process. 

\subsection{Normalization}

ML usually operates under the normalization constraint,
$||z||=1$, where $||. ||$ is a suitable norm, say Euclidean 
for simplicity, i.e. $||z^2|| = \int z^2(q) dq$.  
Within our analogy, such constraint amounts to requiring 
$$\int z(z-z^{eq})dq=0.$$ 
Given the nonlinear structure of $z^{eq}$ it appears pretty unwieldy
to translate the above condition into a constraint on the moments $W_k$.
A better option is to impose the normalization as a 
soft constraint, i.e. by augmenting the dynamics with a damping term
$$
R[z] = -\alpha(||z||-1)
$$
where $\alpha$ measures the coupling of the system with the external
reservoir enforcing the normalization constraint on a time scale $1/\alpha$.
Such terms are commonplace in molecular dynamics and Langevin simulations 
\cite{FRENKEL,META}.

\section{Special cases}

It is of some pedagogical use to consider special instances of the
dynamical rule (\ref{NEU},\ref{ZETA}).

\subsection{Identity: $b \to 0, W \to I, f(z)=z$,} 

In this limit, the ML update reduces to 
$$
\partial_t z = -\gamma(z-z)=0
$$ 
meaning that there is no evolution and the output is the identity $y=x$.

\subsection{Local ODE:  $W \to I$,} 

In this case we obtain:

$$
\partial_t z = -\gamma (z-f(z)+b)
$$ 
This is a local nonlinear ODE with no spatial mixing which converges
in time to the uniform attractor(s) $z^*(b)$, obeying the condition
$f(z^*)-z^* = b$. For instance, with $f(z)=z^2$ and $b=0$ one obtains  
the logistic equation. 
This yields piecewise constant solutions in space, hence very limited
expressive power.  

\subsection{Linear PDE: $f(z)=z$,} 

In this limit we obtain:

$$
\partial_t z = -\gamma (z-Wz+b)
$$ 

This is a linear non-homogeneous PDE, whose spatial structure
depends on the specific nature of the kernel $W$, as discussed in
the previous section. 
The time-asymptotic solution is $z=(W-I)^{-1} b$, $(W-I)^{-1}$ being
the Green function of the weight kernel.
In the inhomogeneous case, the solution often exhibits a rich
structure in space. Yet, being linear, these solutions are invariant
under amplitude rescaling, $z \to \lambda z$, and support linear
superposition. This symmetries greatly simplify their behaviour, thereby
restricting their expressive power as interpolators..
    
\subsection{Neural PDE:}

The full nonlinear and nonlocal ML update corresponds to
a nonlinear "scrambling" of all terms contributing to the linear PDE 
transformation generated by the weight kernel.
In actual fact, the local nonlinear operator performs a selection
based on the amplitude of the signal. For instance, sigmoidal activation
functions, such as $tanh(z)$, implement the idea that inputs below
a given threshold give no output, around the the threshold they output
responds linearly with the input and far above threshold they undergo saturation.
This sort of activation functions are directly inspired to the 
firing activity of actual neurons.
However, modern machine learning often employs qualitatively different
activation functions, such as ReLU, whereby positive signals are left unchanged,
while negative ones are just suppressed (set to zero).      
This nonlinear amplitude-selection filter appears to be crucial 
for the universality of ML interpolators.


\section{Discrete formulation}

Since ML operates on large and yet finite set of discrete data, it is important
to cast the previous formalism in discrete form.
To this purpose, it proves expedient to start by discretizing time.

\subsection{Discrete time marching} 

By marching in time with a simple Euler forward scheme, we obtain:
\begin{equation}
\label{SYSDYN2}
z(t+\Delta t) = -\gamma \; \Delta t(z-f[Z]),
\end{equation}
where $Z=Wz-b$ and indices have been removed for simplicity.
By letting $\omega \equiv \gamma \; \Delta t$, we obtain 
a classical relaxation scheme:
\begin{equation}
\label{ML1}
z(t+\Delta t) = (1-\omega) z + \omega f[Z];
\end{equation}
By evolving this relation over a time span $T=(L+1)\Delta t$, the above scheme
with $\omega=1$ is exactly a ML forward update with $L$ hidden layers and $N$ neurons 
per layer, layers $0$ $(t=0)$ and layer $L$ ($t=T$) corresponding 
to input and output, respectively.

The analogy between machine-learning and dynamical systems is not new, but
the specific identification of local equilibria with the ML target does not
appear to have been highlighted before, nor does the interpretation of the weight
kernel and associated local equilibria in terms of information-propagation processes.

The potential advantages of this interpretation are as follows. 
First, the identification of the
weight kernel as a PDE transformer provides a transparent interpretation of the weight
matrix and consequently of the local equilibrium in terms 
of physical information-propagation processes. 
Second, it suggests more economic strategies based on the optimization of the relevant
momemnts of the weight matrix instead of each of its components.
Third, the relaxation update suggests that $\omega$ may also be 
employed as an effective optimization parameter, possibly a highly relevant one, since
finite-time relaxation is physically related to dissipative effects. 
For instance, this parameter is a crucial ingredient of the 
lattice Boltzmann simulation of fluid flows and other transport phenomena 
\cite{OUP01,RASIN,NATAL}. 

To complete the analogy, we next address space discretization.


\subsection{Space discretization}

In the spirit of handling ML as the evolution of a discrete dynamical system, we
discretize (data) space by first expanding the function $z(q,t)$ onto 
a complete functional basis:
\begin{equation}
z(q,t) = \sum_{i=1}^N z_i(t) \phi_i(q),\;\;\;i=1,N
\end{equation}
where $\phi_i(q) \equiv \phi(q-q_i)$ are suitable local basis functions
centered at $q_i$ and $z_i(t)$ are the associated amplitudes.
Hereafter $q=\lbrace q_1 \dots q_d \rbrace$ denotes a $d$-dimensional array.

By projecting the equation (\ref{ML1}) onto $\phi_j(q)$, we obtain
the following discrete set of equations:
\begin{equation}
\label{DISDYN}
M_{ij} z_j(t+\Delta t) = (1-\omega) M_{ij} z_j(t) + \omega f_i
\end{equation}
where $M_{ij} = \int \phi_i (q) \phi_j(q') dq'$ is the "mass"
matrix and 
\begin{equation}
\label{FI}
f_i = \int \phi_i(q) f[\sum_j \int W(q,q') \phi_j(q') dq' - b(q)]dq
\end{equation}
The latter term is unwieldy, since the nonlinearity does not permit to 
bring $\phi_i(q)$ inside the second integrand to form the double scalar product 
yielding the weight matrix, $W_{ij} = \int W(q,q') \phi_i(q) \phi_j(q') dq dq'$.

This is possible by choosing a singular set of basis functions, 
$\phi_i(q) = \delta(q-q_i)$, in which case we obtain $M_{ij}=\delta_{ij}$ and 
$$
f_i = f(Z_i) = \int \delta(q-q_i) f[\sum_j W(q,q_j)z_j - b(q)] dq
$$
where
$$
Z_i = \sum_j W(q_i,q_j)z_j - b(q_i)
$$
This leads to the identification of the weight and biases simply as the
point-like values of their continuum counterparts, sampled at the data points 
$q_i$ and $q_j$:
$$
W_{ij} = W(q_i,q_j), \;\;\; b_i = b(q_i).
$$
This recovers the standard structure of the ML update, but at the 
expenses of working with a non-differentiable kernel, 
$W(q,q')=\sum_{ij} \delta(q-q_i) W_{ij} \delta (q'-q_j)$.
This is no serious problem, since the kernel transformation previously discussed
does not require any differentiation of the kernel, but only of the signal $z(q)$.
Moreover, it is possible to exploit the local nature of the 
basis functions to perform the integrals via a low-order numerical quadrature 
(a common practice in finite-element computing).
That is:
\begin{equation}
\label{FI2}
f_i = \sum_{k=0}^Q p_{k} \phi_i(q_i+d_k) 
f[\sum_j z_j \sum_{l=0}^Q  W(q_i+d_k,q_j+d_l) p_l \phi(q_j+d_l) - p_k b(q_i+d_k)]
\end{equation}
where $Q$ is the order of the quadrature and $p_k$ are the corresponding weights.
Note that the quadrature nodes are centered about the data points $q_i$, shifted
by the displacements $d_k$, with $d_0=0$. 
This leads to generalized ML update of the form
\begin{equation}
\label{GML}
f_i = \sum_{k=0}^Q \Phi_i^k f[\sum_{l=0}^Q \sum_{j=1}^N W_{ij}^{kl}\Phi_{j}^l z_j - b_{i}^k] 
\end{equation}
where we have set:
$$
\Phi_i^k     = p_k \phi(q_i+d_k),\;
\Phi_j^l     = p_l \phi(q_j+d_l),\;
W_{ij}^{kl}  = W(q_i+d_k,q_j+d_l)
$$
This is more complicated than the usual ML update, as it involves an "inner" 
matrix structure induced by the quadrature indices.
However, since $Q$ is usually a small number, typically $Q<3$ (fifth order accuracy), the actual 
extra-burden is comparatively minor as compared to the potential gain in smoothness. 
Besides, if smoothness is not a priority, even the simplest case $Q=1$ can be used, in 
which case the only  extra-burden is the multiplication by 
the pre-factor $\Phi(q_i)$.
It is therefore speculated that the "matrix" generalization presented by the expression
(\ref{GML}) might offer enhanced accuracy at the prize of an acceptable computational extra-burden.

\section{Sparse, high-dimensional data}

The above formulation is conceptually straightforward and also practically viable
for low-dimensional,space-filling data, for which finite 
elements/volumes/differences provide a computationally efficient representation. 

In this case the forward step of the ML procedure is literally a 
time-integrator of the corresponding integro-differential dynamical system.
As anticipated, the interesting point is the reverse-engineering of the weights
$W_{ij}$ to the moments $W_{ki}$, which delivers a transparent 
physical interpretation of the weight matrix. 

For sparse high-dimensional data, one must turn to cluster techniques,
whereby data are first assigned to a set of clusters \cite{CLU} and the integrals
are replaced by weighted sums of the avalaible data.
More precisely, each point $q_i$ within the given cluster $\Omega_c$ is assigned an equal weight 
$p_i = V_c/N_c$, $i=1,N_c$, where $V_c$ is the volume of the cluster $c$, defined
as the volume of the minimum $d$-dimensional hypercube containing the entire cluster \cite{CLU}.    
The continuum integral can then be replaced by a corresponding weighted sum
$$
f_i =  f[W_{ij} z_j - b_i]
$$
where the factor $V_c/N_c$ has been incorporated in the weights and biases.
 
Cluster integration can be iformally paralleled to a piecewise-constant representation of
the signal, $z(q) = \sum_j z_j \xi(q-q_j)$, where $\xi=1$ if $q$ belongs to the
parcel of cluster assigned to $q_i$ and zero otherwise. 
Since the parcels do not overlap, the mass matrix is diagonal 
but the weight matrix $W_{ij}$ is not, and the corresponding weight 
kernel is given by:
\begin{equation}
\label{CMC}
W(q,q') = \sum_{i,j=1}^{N_c^2} \xi(q-q_i) W_{ij} \xi(q'-q_j)
\end{equation}
where all data belong to the same cluster.

Both finite-grid and cluster formulations show 
that ML operation based on transformers bears a well-defined correspondence
with the evolution of discrete dynamical systems, whose local attractor is
precisely the target of the ML procedure. 

\section{Neural Advection-Diffusion-Reaction equation}

Let us consider the neural equation
$$z_t = -\gamma(z-f[Wz-b]),$$
with a weight kernel corresponding to an advection-diffusion-reaction process:
$$
Z = Wz-b = R z - U z_q + D z_{qq}-b
$$
$D$ being the diffusion coefficient, $U$ the advection speed and $R$ a local
reaction rate, all constant for simplicity.

On a uniform grid with spacing $\Delta $, the discrete set of weights reads 
as follows (biases are set to zero for simplicity)
$$
W_{ij} = A \delta_{i-1,j} + C \delta_{ij} + B \delta_{i+1,j}
$$
where 
$$
A = -U/2\Delta + D/\Delta ^2\; 
B = +U/2\Delta + D/\Delta^2,\; 
C =  1 -2 D/\Delta^2 + R
$$
The moments are $W_0=A+B+C$, $W_1=B-A$, $W_2=A+B$.
All higher order odd moments are equal to $W_1$ and all even ones are equal to $W_2$.
The forward algorihm proceeds in three steps:
\begin{enumerate}
\item{} Compute the transformed signal: $Z_i=A z_{i-1} + C z_i + B z_{i+1}$
\item{} Form the local equilibrium:     $z_i^{eq}=f(Z_i)$
\item{} Advance to next layer:          $z(t+1) = (1-\omega) z + \omega z^{eq}$ 
\end{enumerate}

The backward step proceeds as usual, by minimizing 
the distance of the solution $z(T)$ from the given target $y^T$.
In Euclidean metrics: 
$$
dis(A,B,C;t=T)=\sum_{i=1}^N (z_i(T)-y_i^T)^2
$$
A steepest descent update of the three parameters $A,B,C$, with the gradient
of the above distance computed via the chain rule, completes the first iteration.  

Clearly, there is no guarantee that a generic target can be reached at any
specific time during the evolution of a three-parameter homogeneous ADR 
process such as the one described above.
However, far more freedom can be injected by turning the constant coefficients into
local ones, i,e. $U=U(q)$, $D=D(q)$, $R=R(q)$, thus yielding $3N$ free
parameters, which is still much less that the $O(N^2)$ parameter of a full $W$ matrix.  

An even more aggressive policy is to change the three local parameters "on the fly", i.e.
$U(q,t),D(q,t),R(q,t)$, where the time dependence is steered 
by minimization of the time-dependent distance: 
$$
dis(A,B,C;t)=\sum_{i=1}^N (z_i(t)-y_i^T)^2
$$
This is highly reminiscent of the celebrated Car-Parrinello 
strategy in ab-initio molecular dynamics \cite{CP}. 
It amounts to using $3NL$ parameters, with the benefit that
the chain rule only involves two layers at a time.

It would be interesting to explore the class of targets that can 
be reached by these three progressive families of neural ADR equations.

\subsection{Multi-dimensions}

The multi-dimensional procedure remains the same provided one uses 
the cluster representation of the kernel discussed above.
However, the scalar coefficients now become tensors of rank zero 
(Reaction), one (Advection) and two (Diffusion) 
respectively, with(at most) $1$, $d$ and $d(d+1)/2$ independent components 
respectively in $d$ spatial dimensions.
Even with very large dimensional data, say $d=10^3$, this would still
yield at most one million parameters, much less than present day LLM-based
transformers. For the heterogenous case this gives $O(Nd^2)$ parameters,
comparable to the full matrix case whenever $d \sim N^{1/2}$.
Going to higher-order PDE's, say of order $p$ gives $O(Nd^p)$ parameters,
setting an increasing stringent constraint on the dimensionality, namely
$d \sim N^{1/p}$. 

Given the richness of the patterns generated by the heterogeneous coupling between 
advection, diffusion and reaction mechanisms \cite{NELSON}, it is plausible to 
expect that the neural ADR equations should be well positioned 
to reach a large family of targets with at most $O(Nd^2)$ physically 
explainable parameters.

Future simulation work will tell. 

\section{Summary}

Summarizing, we have highlighted a formal and substantial analogy 
between the forward step of Machine Learning (ML) algorithms and 
discrete dynamical systems in relaxation form.
The analogy identifies the model function of the ML
scheme with the local equilibrium of the discrete
dynamics, thereby offering a transparent interpretation 
of the weights in terms of physical information-propagation processes,
such as advection, diffusion, dispersion and higher order generalizations thereof.
Besides improving the explainability of current ML applications, it is
argued that this analogy may facilitate the development of new explainable ML 
algorithms with a reduced number of weights.

\section{Acknowledgements}

The author is grateful to SISSA for financial support under the
"Collaborations of Excellence" initiative, as well as 
to the Simons Foundation for supporting several enriching visits.
He also wishes to acknowledge many enlightening discussions 
over the years with PV Coveney, M. Durve, A. Laio and D. Spergel.
I am also grateful to R. Natalini for valuable discussions and 
for pointing me to the neural PDE literature.
This paper is dedicated to the memory of Eugenio Beltrami, and all-time master 
of PDEs, on occasion of the delivery of the Beltrami Senior Scientist Prize.

\end{document}